\documentclass[conference]{IEEEtran}
\IEEEoverridecommandlockouts
\usepackage{multirow}
\usepackage{booktabs}
\usepackage{graphicx}
\usepackage{listings}
\usepackage{amsmath}
\usepackage{booktabs}
\usepackage{footnote}

\makesavenoteenv{tabular}
\makesavenoteenv{table}

\usepackage{placeins}
\usepackage{afterpage}
\usepackage{subfigure}
\usepackage{caption}
\usepackage{float}
\usepackage{placeins}
\usepackage{enumitem}
\newcounter{prop}[section]
\renewcommand*{\theprop}{\thesection}

\newcommand{\comment}[1]{}
\makeatletter
\newcommand{\linebreakand}{%
  \end{@IEEEauthorhalign}
  \hfill\mbox{}\par
  \mbox{}\hfill\begin{@IEEEauthorhalign}
}
\makeatother

\usepackage{cite}
\usepackage{amsmath,amssymb,amsfonts}
\usepackage{algorithmic}
\usepackage{graphicx}
\usepackage{textcomp}
\usepackage{xcolor}
\def\BibTeX{{\rm B\kern-.05em{\sc i\kern-.025em b}\kern-.08em
    T\kern-.1667em\lower.7ex\hbox{E}\kern-.125emX}}

\title{Patterns Detection in Glucose Time Series by Domain Transformations and Deep Learning\\

}

\author{\IEEEauthorblockN{J. Alvarado}
\IEEEauthorblockA{
\textit{Universidad de Extremadura}\\
\textit{Universidad Complutense de Madrid}\\
Mérida, Spain \\
jorgal07@ucm.es}
\and
\IEEEauthorblockN{J. Manuel Velasco}
\IEEEauthorblockA{
\textit{Universidad Complutense de Madrid}\\
Madrid, Spain \\
mvelascc@fis.ucm.es }
\and
\IEEEauthorblockN{F. Ch\'avez}
\IEEEauthorblockA{
\textit{Universidad de Extremadura}\\
M\'erida, Spain \\
fchavez@unex.es} 
\linebreakand
\IEEEauthorblockN{J.Ignacio Hidalgo}
\IEEEauthorblockA{
\textit{Universidad Complutense de Madrid}\\
\textit{Instituto de Tecnología del Conocimiento}\\
Madrid, Spain \\
hidalgo@dacya.ucm.es}
\and
\IEEEauthorblockN{F. Fern\'andez de Vega}
\IEEEauthorblockA{
\textit{Universidad de Extremadura}\\
M\'erida, Spain \\
fcofdez@unex.es}
}

\begin{document}



\maketitle

\IEEEpubidadjcol

\begin{abstract}
People with diabetes have to manage their blood glucose level to keep it within an appropriate range. Predicting whether future glucose values will be outside the healthy threshold is of vital importance in order to take corrective actions to avoid potential health damage. In this paper we describe our research with the aim of predicting the future behavior of blood glucose levels, so that hypoglycemic events may be anticipated. The approach of this work is the application of transformation functions on glucose time series, and their use in convolutional neural networks. We have tested our proposed method using real data from 4 different diabetes patients with promising results.
\end{abstract}

\begin{IEEEkeywords}
Diabetes, Glucose Time Series, Deep Learning
\end{IEEEkeywords}

\section{Introduction}

\textit{Diabetes mellitus (DM)} is one of the most relevant diseases in the world and in a few years it will be one of the main causes of death according to the World Health Organisation (WHO). Type I diabetes (T1DM) is an autoimmune disease that affects the pancreas, causing the destruction of the cells responsible for producing insulin, the beta cells. It is important that the pancreas is able to produce insulin to regulate blood glucose levels. Thanks to insulin, the cells absorb the glucose found in the bloodstream. In diabetic patients, the pancreas is unable to supply insulin, which prevents the glucose in the blood from being taken up by the cells. This causes glucose levels to rise to a point that is dangerous to health, known as hyperglycemia. As a result, diabetic patients have to replace the pancreas' natural function, the production of insulin, by means of injections or so-called \textit{insulin boluses}. Controlling the exact amount and type of insulin is a critical task, as a mistake can cause the glucose level to reach dangerous levels. It is important to keep blood glucose levels in a healthy range to avoid hypoglycemic and hypoglycemic events that can cause significant damage to health, which, if prolonged over time, can even lead to death. Estimating a patient's glucose behaviour in the immediate future is challenging, as numerous factors can intervene to change healthy glucose levels, even suddenly. To determine the future value, it is essential to know the current glucose level and this is obtained using devices called continuous glucose monitors (CGMs), which consist of a sensor that measures the glucose value in time intervals. But the analysis of this factor is not the determining factor in establishing an optimal glucose level; there is the added difficulty of estimating the grams of carbohydrates ingested since, depending on the type and quantity of these, they can affect glucose level in one way or another. The temporal effects of different types of insulin boluses also need to be taken into account, as well as other factors such as exercise, age, stress, etc.

The motivation for this work stems from the need to create predictive models to determine, in patients with type 1 diabetes, the future behaviour of blood glucose with the aim of keeping it in a healthy range. If an accurate prediction of a patient's future glucose values can be obtained, the patient will be able to make decisions and take corrective measures to prevent glucose reaching a value that could cause potential long-term damage to health. Throughout this paper we present a preliminary study of glucose value prediction using only the values obtained from glucose meters.

As mentioned above, the glucose time series of each patient is used to predict future glucose values, and whether these may be detrimental to the patient's health. To do this, a transformation of the time series is performed by means of transformation functions, which will allow us to obtain an image that represents the glucose values over time. These resulting images are used to train predictive models with Convolutional Neural Networks (CNN) to detect patterns in glucose and predict the future behaviour of blood glucose levels.

The paper is organized as follows. Section \ref{sec.glucose.prediction} explains in more detail the problem of determining the future behavior of glucose in diabetic patients and revises other approximations to the problems recently presented. In section \ref{sec:state_of_the_art}, we provide a brief overview of our work within the framework of the current state of the art. Section \ref{methodology} explains our approach, the workflow and the main techniques used in this research, whereas Section \ref{sec:experimental} describes the experimental setup and the forecasting results that these techniques have obtained. Conclusions and future work are exposed in Section \ref{sec:conclusion}.

\section{Glucose Prediction}
\label{sec.glucose.prediction}

As mentioned in the introduction, in patients with type 1 diabetes or insulin-dependent, the disease causes the destruction of the so-called beta cells found in the pancreas. In a healthy pancreas, these cells are responsible for producing the insulin that is distributed into the bloodstream. Insulin is of vital importance because it is responsible for regulating the glucose level in the blood, allowing the cells to absorb it from the blood. In the case of a person with diabetes, the pancreas does not produce insulin which prevents the assimilation of glucose by the cells, causing an increase in blood glucose levels dangerous to health called hyperglycemia. 

Managing blood glucose levels is a challenge that people with diabetes usually have to perform every day in order to keep them within a healthy range. In the case of patients with type 1 diabetes, they must estimate the effects insulin can affect their blood glucose. This task is usually performed manually based on:

\begin{itemize}
   \item Current glucose level: this value can be obtained manually with a glucometer or with a continuous glucose monitor (CGM).
   \item Carbohydrate intake: estimation of the amount ingested in grams. Depending on the amount and type of food it can have different effects on blood glucose.
   \item Insulin injections: the amount and type of insulin boluses to be injected should be calculated based on current glucose and grams of carbohydrates ingested.
   \item Physical activity: deciding whether to do more or less exercise and choosing the right intensity.
\end{itemize}

Correct estimation of the glucose level is essential to avoid episodes of hyperglycemia and hypoglycemia. Hyperglycemia occurs when the glucose value is $> 180  mg/dL\ (10 mmol/L)$ and causes damage to multiple organs, tissues, blood vessels and nerves if maintained for a long period of time. On the opposite side is hypoglycemia, manifested when the blood glucose level is $< 70 mg/dL\ (5.5 mmol/L)$. 

Hypoglycemia is a complication of diabetes and has much more severe long-term effects with respect to hyperglycemia because it increases the risk of cardiovascular disease, increase the risk of kidney disease, affects the nervous system, etc., among others. Severe hypoglycemia can impair cognitive function, producing microvascular damage to the brain and even death. 

Episodes of hypoglycemia can occur at any time of the day but are usually classified into daytime, postprandial and nocturnal hypoglycemia. Especially the latter, hypoglycemia events during the night are really dangerous because they usually go unnoticed due to sleep. As they are symptomatic, the effects are prolonged, seriously affecting the patient's health and in the worst cases even leading to the dead-in-bed syndrome.


The main objective of this work is the creation of predictive models for the early detection of hypoglycemia episodes. Accurate prediction would be of great help to diabetic patients because it would allow them to take timely decisions and actions to correct their low glucose levels. This would produce great benefits in many aspects of the patient's life such as a considerable improvement in their health and can even save lives.

\section{State of the Art}\label{sec:state_of_the_art}

Several research works have used different mathematical transformations as a preliminary step for the classification of time series. One of the papers that initiated this field of experimentation is the one by Ye et al. \cite{Ye2009TimeSS}, together with Hsu et al. \cite{HSU20152838}. The former sought to find specific particularities of the time series while the latter sought to discard part of the series in order to find common patterns of dynamic behavior that would allow them to be grouped together. 
Before the advent of deep learning in this field, we have the paper by Wang et al. \cite{Wang2012ExperimentalCO} that summarizes the techniques used to classify time series up to that time. And in Ji et al. \cite{Ji2018A2T}, we have a posterior equivalent description of transform techniques (without including Wavelets).

Researchers in the financial world were probably the most enthusiastic advocates of applying the wavelet transform to time series. As an example we have the work of Masset (2015) \cite{Masset2015} and Bolman et al. \cite{Bolman19Morlet}. This research has been extended to other time series in Grant and Islam \cite{Grant2021SignalCU}.

Recently, we can find examples of researchers concerned with the classification of multivariate time series, Pasos et al.  \cite{DBLP:journals/datamine/RuizFLMB21} and Middlehurst et al. \cite{DBLP:journals/corr/abs-2104-07551}. The second one studies the inclusion of an ensemble of classifiers, an option that could also be used for univariate time series. 

So our work can be considered an extension or continuation of these works but with the objective of predicting hypoglycemia scenarios within a one-day horizon.
Within the research literature focused on the prediction of hypoglycemia events we can find several examples using various machine learning techniques and mostly with prediction horizons ranging from 30 minutes to 4 hours.

Examples of shorter-term horizons can be found in the works of Dave et al. \cite{dave2021feature} and in \cite{quan2019ai}. The former uses a mixture of various machine learning techniques, while the latter applies a recurrent neural network, specifically a long-short term memory network. Another similar example that uses 4 different machine learning techniques and is focused on short horizons immediately after meals is that of Seo et al. \cite{seo2019machine}.

Among the works focused on longer prediction horizons, we can cite Oviedo et al. \cite{oviedo2019risk} which uses support vector machines and seeks to predict hypoglycemia events in the four hours following a meal. 

Given that nocturnal hypoglycemia is particularly dangerous as it can lead to sudden death (dead-in-bed syndrome), we can find research studies specialized in its prediction, such as the examples of Vu et al. \cite{vu2019predicting} and Porumb et al. \cite{porumb2020precision}. The first one uses random forest and divides the night into three periods, obtaining the best prediction results in the period closest to dinner. The second has a hybrid approach mixing convolutional and recurrent networks and has a special place in the literature for making use of electrocardiogram time series.

\section{Methodology}\label{methodology}

\begin{figure}[!ht]
\centering
\includegraphics[width=0.45\textwidth]{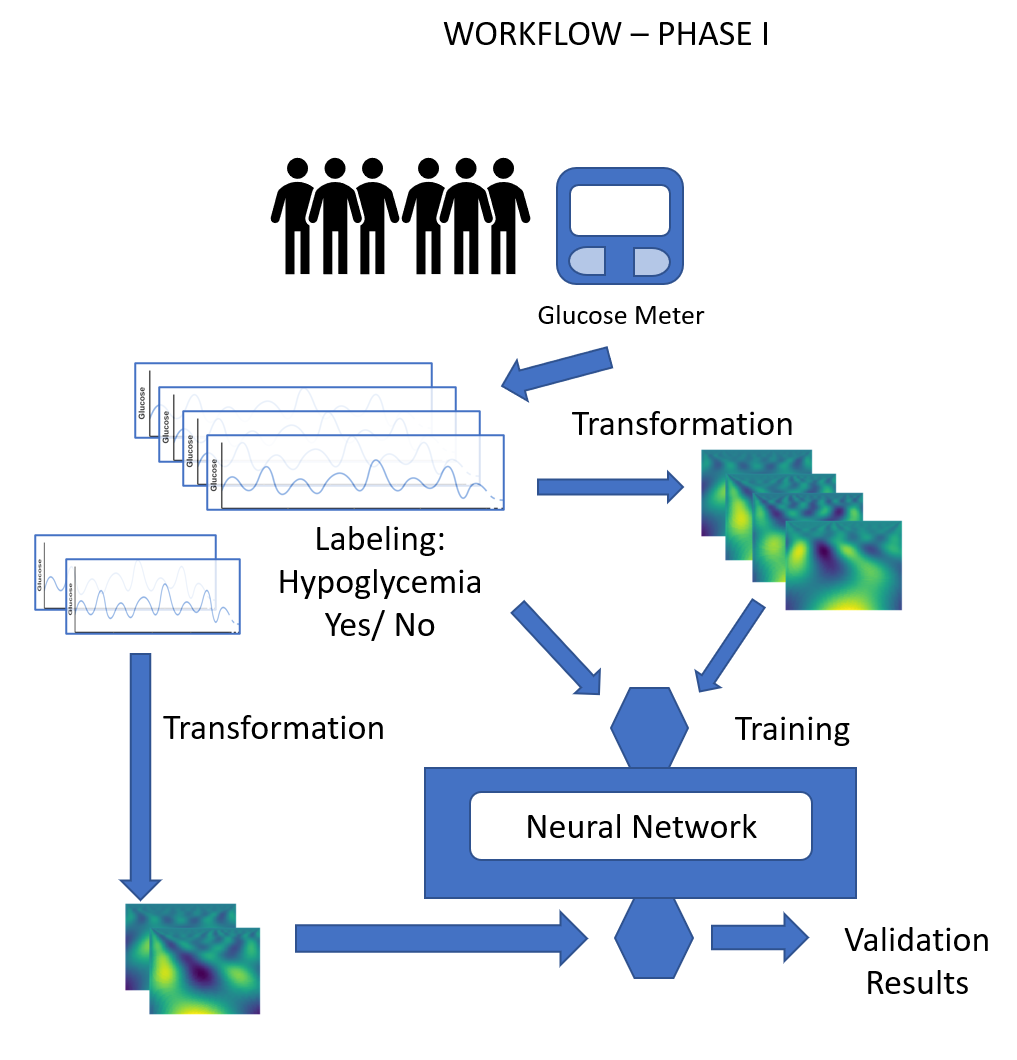}
\caption{Workflow. Phase I. Training/ Validation}
\label{fig:workflow1}
\end{figure}

We can see in figure \ref{fig:workflow1} the first phase of our workflow. We have developed a dataset from data collected using a continuous sensor that records the blood glucose levels of several people as a time series (Subsection \ref{subsec:glucose_time_series}). These time series comprise several days and have been segmented and labeled depending on whether the person suffered hypoglycemia on the subsequent day. This dataset has been also enriched using a data augmentation technique (Subsection \ref{subsec:data_augmentation}). Part of this dataset is used to train the model and part is used for the validation subphase.

Finally, before {\it training} the neural network, the glucose time series undergoes a multilevel non-linear transformation that allows the same information to be expressed as an image (\ref{subsec:transform_functions}). With these images we train a Convolutional Neural Network (subsection \ref{subsec:cnn}) and obtain the model that we will use in the {\it test} phase. The neural network is trained to classify the images according to their label, i.e., whether the patient suffered hypoglycemia or not on the posterior day. 

\begin{figure}[!ht]
\centering
\includegraphics[width=0.35\textwidth]{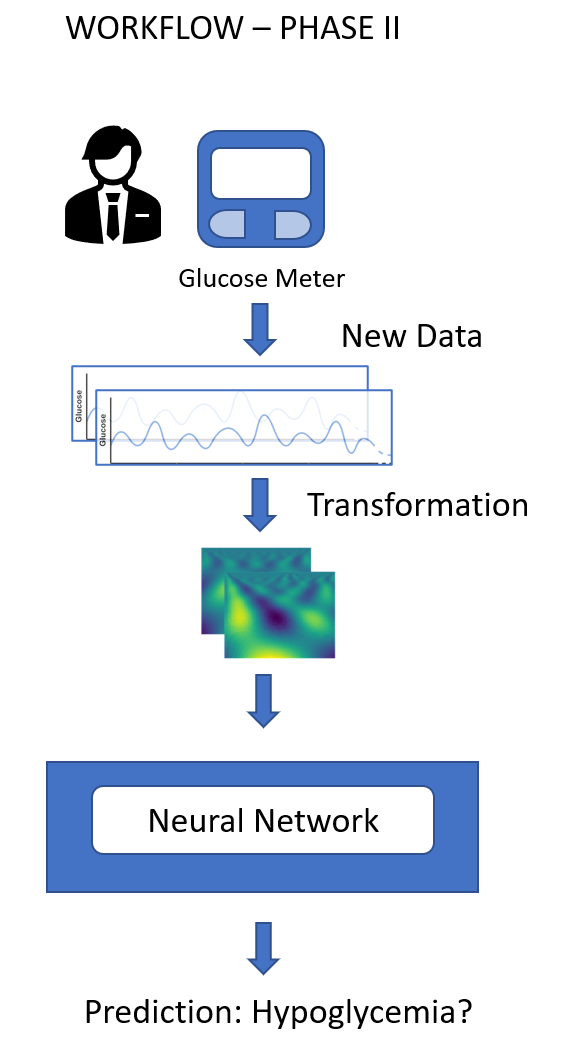}
\caption{Workflow. Phase II. Prediction}
\label{fig:workflow2}
\end{figure}

In the test phase, Figure \ref{fig:workflow2}, new data is entered into the model which provides a classification that will indicate whether or not hypoglycemia occurs. We will use this classification as a way to predict possible hypoglycemia on the next day.

\subsection{Glucose Time Series}
\label{subsec:glucose_time_series}



Blood glucose data were collected using a continuous glucose monitoring (CGM) system. These systems are widely used today as they offer numerous advantages over manual glucometers. One of the main advantages is that they automatically measure glucose at five to fifteen minute intervals using a sensor. 

With the data obtained by the GCM, the glucose evolution of a given patient can be observed. This data allows us to work with a patient's glucose time series, which contains data from different days. The series obtained is a two-dimensional series in which the x-axis represents the time at which the data was taken and the y-axis represents the glucose value obtained by the CGM. The series will be continuous over time, provided that the CGM is working properly. A continuous measurement over several days will present a complete history of the patient's glucose evolution and any episodes of hypoglycaemia or hyperglycaemia. These detected episodes will help us to predict future abnormal events for the patient.

\subsection{Application of transform functions to time series}
\label{subsec:transform_functions}

The next step in our approach is to prepare the data so that they can be used for training models with CNNs. These networks require the input to be an image and the data collected from patients are numerical glucose values over time, and it is necessary to convert this information to images. For this purpose, transformation functions have been applied to the glucose time series with the objective of generating images that include the patterns of glucose behavior.

\begin{figure}[t]
\centering
\begin{tabular}{c}
\includegraphics[width=0.45\textwidth]{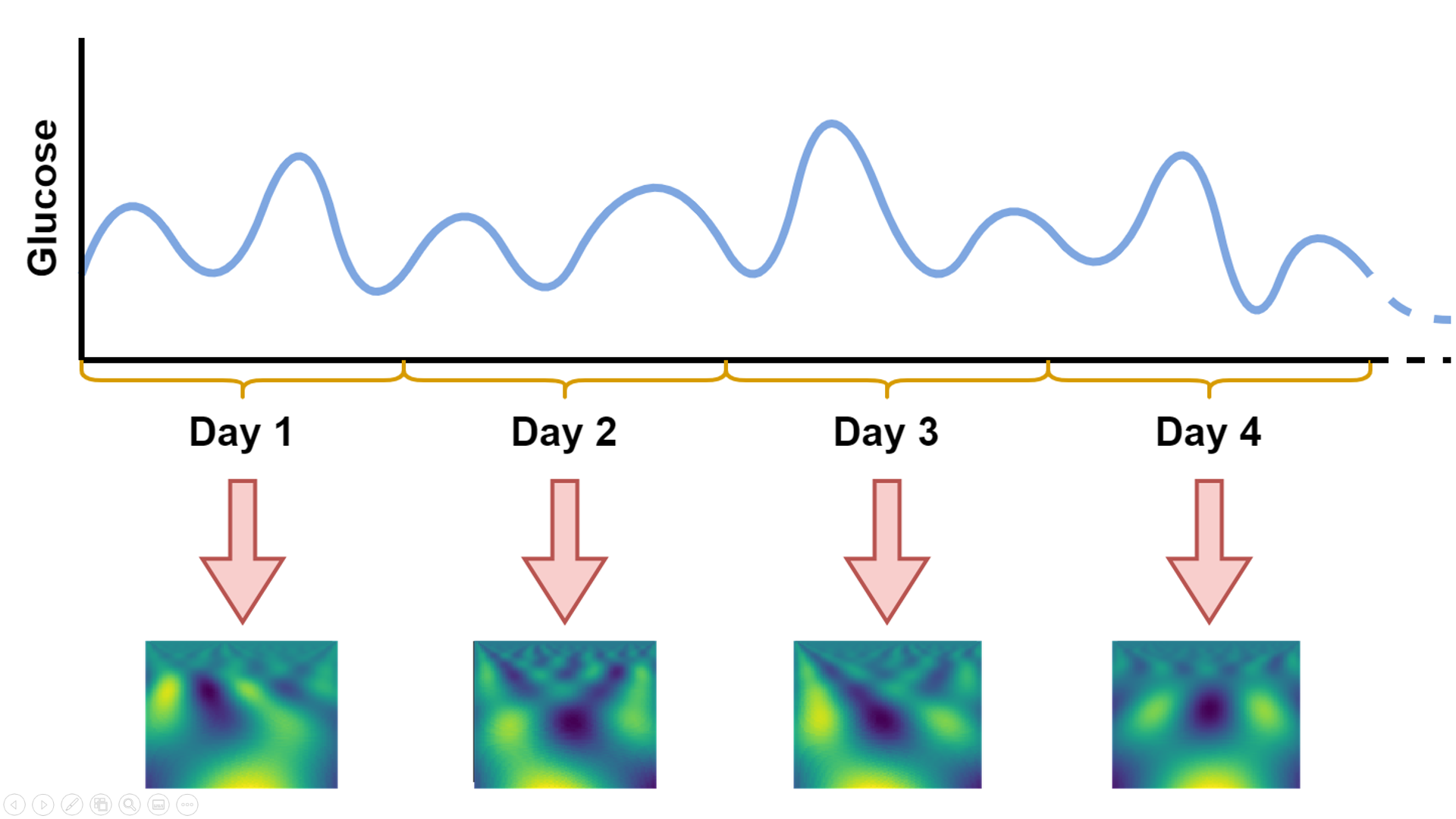}
\end{tabular}
\caption{Application of transform functions to glucose time series}
\label{fig:transform_functions}
\end{figure}


As can be seen in Figure~\ref{fig:transform_functions} the time series is divided into days. By applying a transformation function, an image is obtained that will represent the glucose measurements of the analyzed day (24 hours).

For a CNN to be optimized, a set of images is necessary, and these images must be labeled in different classes. In this way, an image classification model is obtained for a certain problem. The objective of the work presented in this paper is to obtain a predictive model of glucose levels, based on the time series data of the values obtained by the CGM. As we can see we obtain an image that must be labeled as hypoglycemic or normal. In order to label the image we must look at the values of the following 24 hours after the image is obtained. If in these values we find an episode of hypoglycemia, $glucose\ value < 70\ mg/dL\ (5.5\ mmol/L)$, the image will be labeled as hypoglycemia, otherwise it will be labeled as normal.

The model obtained based on CNNs in the work presented here can be considered as a predictive model, although actually the CNN optimization process is a classification. We consider it a predictive model, since the labeling of the images has been done considering if a hypoglycemia episode occurs in the following 24 hours after the image has been created, so the system will be able to detect abnormal glucose ranges in the following 24 hours after the image has been taken.

\subsection{Data augmentation}
\label{subsec:data_augmentation}


Considering that our dataset is limited, the number of images obtained from the process described in the previous section is quite small. With a reduced dataset the predictive models cannot be optimised correctly, for a correct training of the models it is necessary to obtain a larger dataset, therefore, we must use some technique that allows us, with the patient's glucose dataset and without using synthetic data, to increase the dataset.

\begin{figure}[t]
\centering
\begin{tabular}{c}
\includegraphics[width=0.48\textwidth]{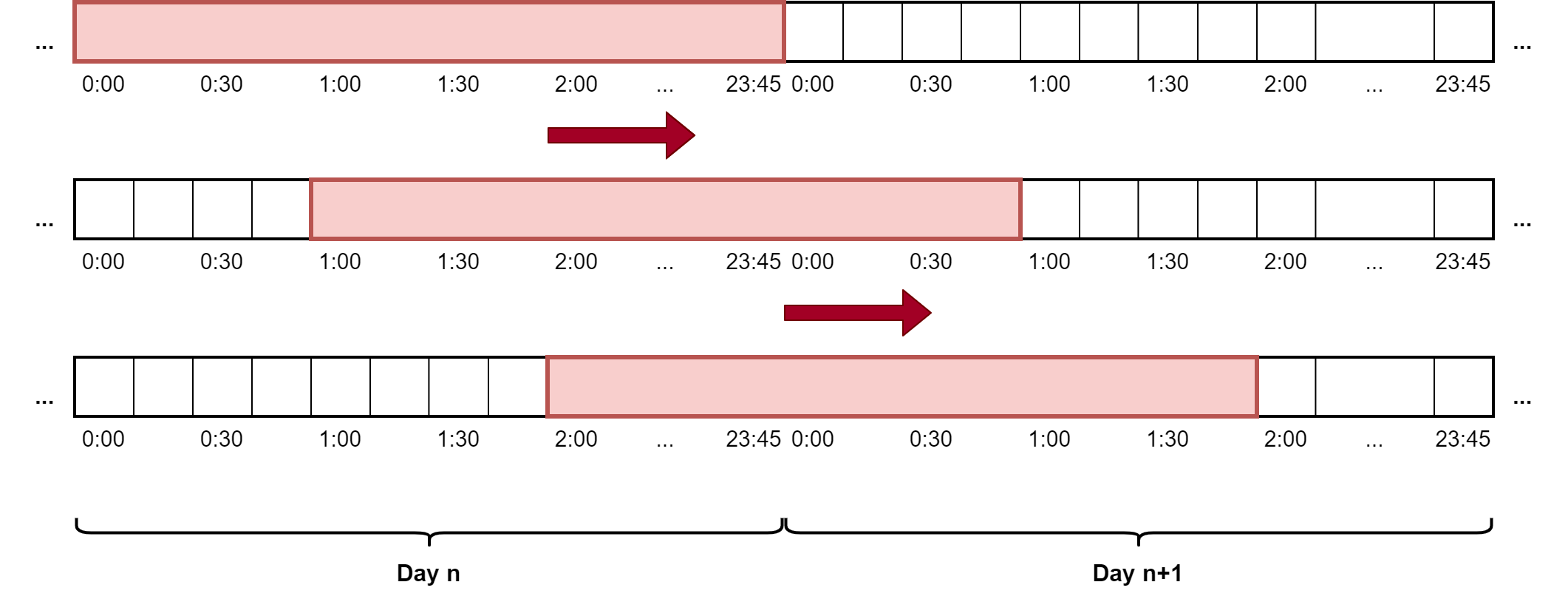}
\end{tabular}
\caption{Data augmentation using a rolling window of 1 hour size}
\label{fig:dataaugmentation}
\end{figure}


The figure~\ref{fig:dataaugmentation} shows graphically the technique used in the dataset augmentation process. It is a sliding window with a 1-hour step. As explained in the previous section, a 24-hour set of the patient's glucose data is used to obtain, through a transformation, an image that characterises that period. As indicated in the labelling process, the next 24 hours are used to determine the category to which the image belongs. In order to obtain the largest number of images, the 24-hour window is moved with a step of 1 hour, thus obtaining the following values within the window, and their corresponding image by means of the transformation. The process of labelling this new image is similar to the one detailed above. 

This technique has allowed us to have a larger dataset, but it also allows us to describe a much more significant evolution of the patient's glucose, by analysing its evolution in 24-hour windows shifted with a 1-hour step.

\subsection{Convolutional Neural Networks}
\label{subsec:cnn}

Convolutional neural networks (CNN) are a type of neural network used in computer vision that simulate the functioning of neurons in the primary visual cortex. These networks are used for object detection, classification, anomaly detection, time series based prediction among others.

\begin{figure*}[t]
\centering
\begin{tabular}{c}
\includegraphics[width=0.99\textwidth]{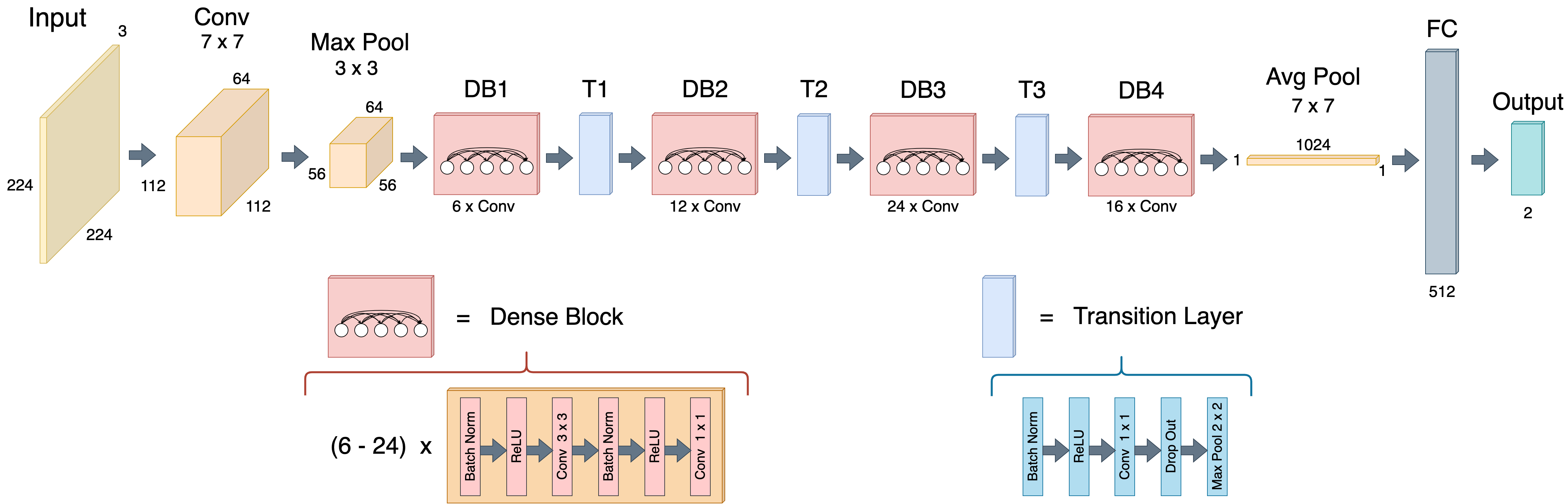}
\end{tabular}
\caption{DenseNet-121 Architecture }
\label{fig:densenet121}
\end{figure*}

Multiple CNN architectures have been developed in recent years, but in this paper we will focus on DenseNet \cite{huang2017densely} networks. This architecture has the advantage of simplifying the pattern of connectivity between layers ensuring maximum information flow and requiring fewer parameters with respect to traditional CNN architectures. There are different DenseNet network configurations that are characterized by the number of convolution layers and feature maps. In particular, DenseNet-121 architecture has been used in this work.

Figure~\ref{fig:densenet121} shows the structure of DenseNet-121.  First, the network has as input RGB images of size 224 x 224 pixels and then a 7 x 7 convolution is applied. Then, DenseNet is organized in dense blocks and transition layers are placed between these blocks. A dense block is the combination of several layers formed by the composite function of batch normalization, ReLU and 3x3 or 1x1 convolution. These layers are interconnected with each other so that the produced feature maps are concatenated at the end of the block adding new information that can be reused. The transition layers are responsible for subsampling the feature maps between the dense blocks by batch normalization following a 1 x 1 convolution and finally a 2 x 2 max pooling. At the end of the last dense block an average pooling of 7 x 7 is applied and the output is connected to a dense layer of size 512. Finally, the classification layer has been adapted to our problem by using a layer of size 2 with softmax function.


\section{Experimental Results}
\label{sec:experimental}

\subsection{Experimental set-up}
\label{subsec:setup}

In order to run the experiments, we used a computer with an Intel Core i7-6850K @ 3.6 GHz, 64 GB RAM and two GPUs: NVIDIA GeForce GTX 1080 Ti with 11 GB VRAM and NVIDIA GeForce GTX 1080 with 8 GB VRAM. To perform the training of the models, the python Tensorflow library has been used together with the pre-trained DenseNet-121 network provided by the Keras library and modified to adapt it to our problem as mentioned in section~\ref{subsec:cnn}.


A dataset of 866 images from four patients was used. For the training set 75 \% of the images have been used, for the validation set 15 \% and for the test set 10 \%. The training of the models has been run 10 times using the parameters detailed in table~\ref{tab:trainingparams}. 
Each model training required an average of 84.89 seconds to complete.

\begin{table}[h]
\centering
\begin{tabular}{|l|l|}
\hline
\multicolumn{1}{|c|}{\textbf{Param}} & \multicolumn{1}{c|}{\textbf{Value}} \\ \hline
Epochs                               & 50                                  \\
Base Learning Rate                   & 0.001                               \\
Image Size                           & 224 x 224 x 3                       \\
Batch Size                           & 64                                  \\
Optimizer                            & Adam                                \\
Loss Function                        & Categorical Crossentropy            \\ \hline
\end{tabular}
\caption{Training Params}
\label{tab:trainingparams}
\end{table}

\subsection{Data}
\label{subsec:data}

It is important to mention that we have used our data set, collected through collaboration with patients and medical staff at the Hospital Príncipe de Asturias in Alcalá de Henares, Spain. The hospital ethics committee approved data collection for our studies.

As in other studies in this field, the collection of these data is not an easy task due to different factors, which means that the datasets are usually small. First of all, sensitive information from patients is being processed. This collection must involve both medical staff and patients who must wear devices for this purpose for several days. Information often has to be discarded because of patient mistakes or device failures.

Data from four patients with Type 1 Diabetes Mellitus (T1DM) were used. Data were collected using the FreeStyle Libre Continuous Glucose Monitoring System (CGM) at fifteen-minute intervals for two consecutive weeks. The patient characteristics is female (50 \%), average age 35.78 (+/- 13.04), weight 69.73 (+/- 19.11) kg, years of disease 12.64 (+/- 8.49), HbA1c average of 7.68 \% (+/- 0.65 \%).

\subsection{Results}
\label{subsec:results}

\begin{figure}[!ht]
\centering
\includegraphics[width=0.45\textwidth]{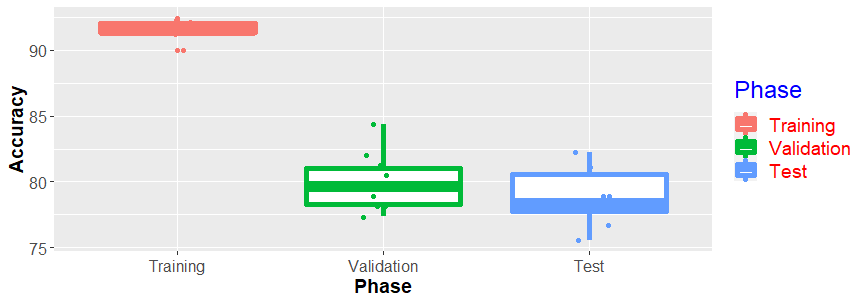}
\caption{Phases Comparison}
\label{fig:boxplot_phases}
\end{figure}

Table \ref{tab:general_results} shows the accuracy obtained in the different models optimized, in the training, validation and test phase. The last two rows correspond to the average and standard deviation of the 10 models. It can be seen that the predictive models obtained have an accuracy greater than 91 \% in training and 78 \% in testing. The box plot of the accuracy obtained in the different phases is shown in figure \ref{fig:boxplot_phases}. After looking at this figure, a question arises: is there a clear difference between the validation group and the test group? To answer this question we will use a statistical analysis \cite{sheskin2007handbook}:

\begin{enumerate}
    \item To be sure that the two groups come from a Gaussian distribution we will use the Saphiro-Wilk test. Here, the Null hypothesis is that the samples are normally distributed. This test yields a p-value of 0.5 for the two groups, so we cannot reject the Null hypothesis.    
    \item Next, we us a F test to compare the variances of the two groups, being the Null Hypothesis that the true ratio of variances is equal to 1. The p-value is bigger than 0.9 so and therefore, we can, now, use an unpaired t-test.
    \item The Null Hypothesis of the t test is that the means of the two groups are not significantly different. The p-value is 0.2, i.e., we cannot reject the Null hypothesis. 
\end{enumerate}

The statistical analysis indicates that there are no significant differences between the means of the validation group and the test group. We can interpret this in two ways:

\begin{itemize}
    \item Our model does not suffer from overfitting.
    \item Performance during validation phase may be a good indicator of future results in a real-world test scenario.
\end{itemize}

\begin{table}[h]
\centering
\begin{tabular}{|c|c|c|c|}
\hline
\textbf{Model} & \textbf{Training} & \textbf{Validation} & \textbf{Test} \\ \hline
1         & 92.13             & 78.13               & 82.22         \\
2         & 91.36             & 82.03               & 75.56         \\
3         & 92.44             & 81.25               & 81.11         \\
4         & 91.20             & 77.34               & 81.11         \\
5         & 91.36             & 80.47               & 78.89         \\
6         & 92.28             & 78.91               & 77.78         \\
7         & 91.51             & 79.69               & 77.78         \\
8         & 91.82             & 78.13               & 76.67         \\
9         & 91.51             & 84.38               & 78.89         \\
10        & 89.97             & 79.69               & 77.78         \\ \hline
Avg       & 91.56             & 80.00               & 78.78         \\
Std      & 0.67              & 2.01                & 2.02          \\ \hline
\end{tabular}
\caption{Results of the different models}
\label{tab:general_results}
\end{table}

\begin{table}[h]
\centering
\begin{tabular}{|c|c|c|c|c|}
\hline
\textbf{Model} & \textbf{Patient 1} & \textbf{Patient 2} & \textbf{Patient 3} & \textbf{Patient 4} \\ \hline
1              & 76.92              & 66.67              & 82.14              & 95.83              \\
2              & 76.92              & 58.33              & 75.00              & 83.33              \\
3              & 69.23              & 83.33              & 85.71              & 87.50              \\
4              & 73.08              & 66.67              & 78.57              & 100.00             \\
5              & 76.92              & 91.67              & 67.86              & 87.50              \\
6              & 73.08              & 75.00              & 78.57              & 83.33              \\
7              & 76.92              & 83.33              & 75.00              & 79.17              \\
8              & 69.23              & 75.00              & 75.00              & 87.50              \\
9              & 65.38              & 83.33              & 82.14              & 87.50              \\
10             & 69.23              & 66.67              & 82.14              & 87.50              \\ \hline
Avg            & 72.69              & 75.00              & 78.21              & 87.92              \\
Std            & 4.02               & 9.86               & 4.91               & 5.73               \\ \hline
\end{tabular}
\caption{Results of applying the test set for each patient to the models}
\label{tab:patients_test_result}
\end{table}

\begin{figure}[!ht]
\centering
\includegraphics[width=0.45\textwidth]{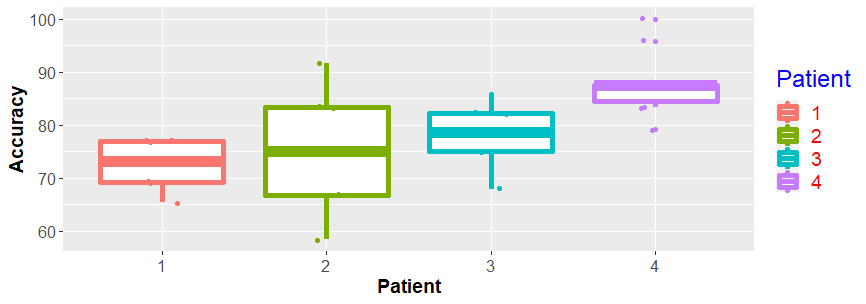}
\caption{Patients Comparison}
\label{fig:boxplot_pacientes}
\end{figure}

The accuracy obtained using the test set for each patient for all models are shown in table \ref{tab:patients_test_result}. The last two rows are the average and standard deviation of the 10 models for each patient. Because the characteristics of each patient are totally different, the results vary but it can be seen that the predictive models obtain up to 88 \% accuracy in the best case and 73 \% in the worst case. Figure \ref{fig:boxplot_pacientes} shows a box plot as a comparison of the accuracy between patients.

We cannot forget that this is a preliminary study with the intention of testing this approach. But as we have seen, the experimental results encourage us to continue investigating this methodology, which seems to deserve to be studied in depth.

\section{Conclusions}
\label{sec:conclusion}

In this work we have investigated the performance of applying transformation functions to glucose time series and using them for training predictive models with CNNs. This is a novel technique that has not been applied before to our knowledge. This technique can predict hypoglycemia events over the course of a day, using only the time series of blood glucose levels from the previous day. This has the main benefit of improving the lives of diabetic patients since, thanks to this technique, they can make early decisions in order to avoid dangerous situations for their health. According to the experimental results obtained, the predictive models offer good performance with high accuracy rates which encourages us to continue investigating this line in the future.





\section*{Acknowledgments}

This work has been  supported by Madrid Regional Goverment and  FEDER funds under grant Y2018/NMT-4668 (Micro-Stress- MAP-CM). We acknowledge support from Spanish Ministry of Economy and Competitiveness under project RTI2018-095180-B-I00 and PID2020-115570GB-C21, PID2021-125549OB-I00  Madrid Regional Goverment - FEDER grant B2017/BMD3773 (GenObIA-CM), Junta de Extremadura, Consejer\'ia de Econom\'ia e Infraestructuras, del Fondo Europeo de Desarrollo Regional, ”Una manera de hacer Europa”, project GR21108. Devices for adquiring data from patients were adquired under the support of Fundaci\'on Eugenio Rodriguez Pascual 2019 grant - \textit{Desarrollo de sistemas adaptativos y bioinspirados para el control gluc\'emico con infusores subcut\'aneos continuos de insulina y monitores continuos de glucosa  (Development of adaptive and bioinspired systems for glycaemic control with continuous subcutaneous insulin infusors and continuous glucose monitors)}.

\bibliographystyle{unsrt} 
\bibliography{arxiv_wavelets} 

\end{document}